\documentclass[11pt,a4paper]{article}
\usepackage[hyperref]{emnlp-ijcnlp-2019}
\usepackage{times}
\usepackage{latexsym}

\usepackage{url}

\usepackage{graphicx}
\usepackage{amsmath}
\usepackage{amssymb}

\usepackage{multirow}
\usepackage{enumitem}
\usepackage{algorithm}
\usepackage{stfloats}
\usepackage{booktabs}
\usepackage{colortbl}
\usepackage{subcaption}
\usepackage{array}   
\usepackage{graphics}
\usepackage{arydshln}

\usepackage{calc}
\usepackage{comment}
\usepackage{wrapfig}

\newcommand{\RN}[1]{%
	\textup{\lowercase\expandafter{\it \romannumeral#1}}%
}

\newcommand{\eg}[0]{\emph{e.g., }}

\newcommand{\beq}{\vspace{0mm}\begin{equation}}
\newcommand{\eeq}{\vspace{0mm}\end{equation}}
\newcommand{\beqs}{\vspace{0mm}\begin{eqnarray}}
\newcommand{\eeqs}{\vspace{0mm}\end{eqnarray}}
\newcommand{\barr}{\begin{array}}
\newcommand{\earr}{\end{array}}

\newcommand{\Wmat}[0]{{{\bf W}}}

\newcommand{\av}[0]{{\boldsymbol{a}}}

\newcommand{\cv}[0]{{\boldsymbol{c}}}

\newcommand{\ev}[0]{{\boldsymbol{e}}}

\newcommand{\hv}[0]{{\boldsymbol{h}}}

\newcommand{\sv}[0]{{\boldsymbol{s}}}

\newcommand{\xv}{\boldsymbol{x}}

\newcommand{\zv}{\boldsymbol{z}}

\newcommand{\thetav}{\boldsymbol{\theta}}

\newcommand{\tauv}[0]{{\boldsymbol{\tau}}}

\newcommand{\Xcal}{\mathcal{X}}

\newcommand{\Dcal}{\mathcal{D}}

\usepackage{color, colortbl}
\definecolor{Gray}{gray}{0.93}

\aclfinalcopy 

\newcommand{\short}{\textsc{PreSS}}
\newcommand{\longname}{\textsc{\textbf{Pre}trained LMs and \textbf{S}tochastic \textbf{S}ampling}}
\newcommand{\bertbase}{BERT}

\newcommand{\gpt}{GPT}

\newlength\myheight
\newlength\mydepth
\settototalheight\myheight{Xygp}
\newcommand*\inlinegraphics[1]{%
  \settototalheight\myheight{Xygp}%
  \settodepth\mydepth{Xygp}%
  \raisebox{-\mydepth}{\includegraphics[height=\myheight]{#1}}%
}

\title{Robust Navigation with Language Pretraining and Stochastic Sampling}

\author{Xiujun Li\textsuperscript{$\spadesuit\diamondsuit$}\quad Chunyuan Li\textsuperscript{$\diamondsuit$}\quad Qiaolin Xia\textsuperscript{$\clubsuit$}\quad Yonatan Bisk\textsuperscript{$\spadesuit\diamondsuit\heartsuit$}\quad\\
\bf{Asli Celikyilmaz}\textsuperscript{$\diamondsuit$}\quad \bf{Jianfeng Gao}\textsuperscript{$\diamondsuit$}\quad \bf{Noah A. Smith}\textsuperscript{$\spadesuit\heartsuit$}\quad \bf{Yejin Choi}\textsuperscript{$\spadesuit\heartsuit$} \\
\textsuperscript{$\spadesuit$}Paul G. Allen School of Computer Science \& Engineering, University of Washington\\
\textsuperscript{$\clubsuit$}Peking University\quad \textsuperscript{$\diamondsuit$}Microsoft Research AI\quad \textsuperscript{$\heartsuit$}Allen Institute for Artificial Intelligence\\
{\tt \{xiujun,ybisk,nasmith,yejin\}@cs.washington.edu}\\
{\tt xql@pku.edu.cn\quad \{xiul,chunyl,jfgao\}@microsoft.com}
}

\date{}

\begin{document}
\maketitle
\begin{abstract}
Core to the vision-and-language navigation (VLN) challenge is building robust instruction representations and action decoding schemes, which can generalize well to previously unseen instructions and environments. 
In this paper, we report two simple but highly effective methods to address these challenges and lead to a new state-of-the-art performance. First, we adapt large-scale pretrained language models to learn text representations that generalize better to previously unseen instructions. Second, we propose a stochastic sampling scheme to reduce the considerable gap between the expert actions in training and sampled actions in test, so that the agent can learn to correct its own mistakes during long sequential action decoding. 
Combining the two techniques, we achieve a new state of the art on the Room-to-Room benchmark with 6\% absolute gain over the previous best result (47\% $\rightarrow$ 53\%) on the {\em Success Rate weighted by Path Length} metric.

\end{abstract}


\section{Introduction}



The vision-and-language navigation (VLN) task, learning to navigate in visual environments based on natural language instructions, has attracted interest throughout the artificial intelligence research community~\cite{hemachandra2015learning,anderson2018vision,chen2018touchdown,savva2019habitat}. It fosters research on multimodal representations and reinforcement learning, and serves as a test bed for many real-world applications such as in-home robots.


\begin{figure}[ht!]
	\vspace{-0mm}
	\centering
	\begin{tabular}{c}
		\hspace{-2mm}
		\includegraphics[width=0.98\columnwidth]{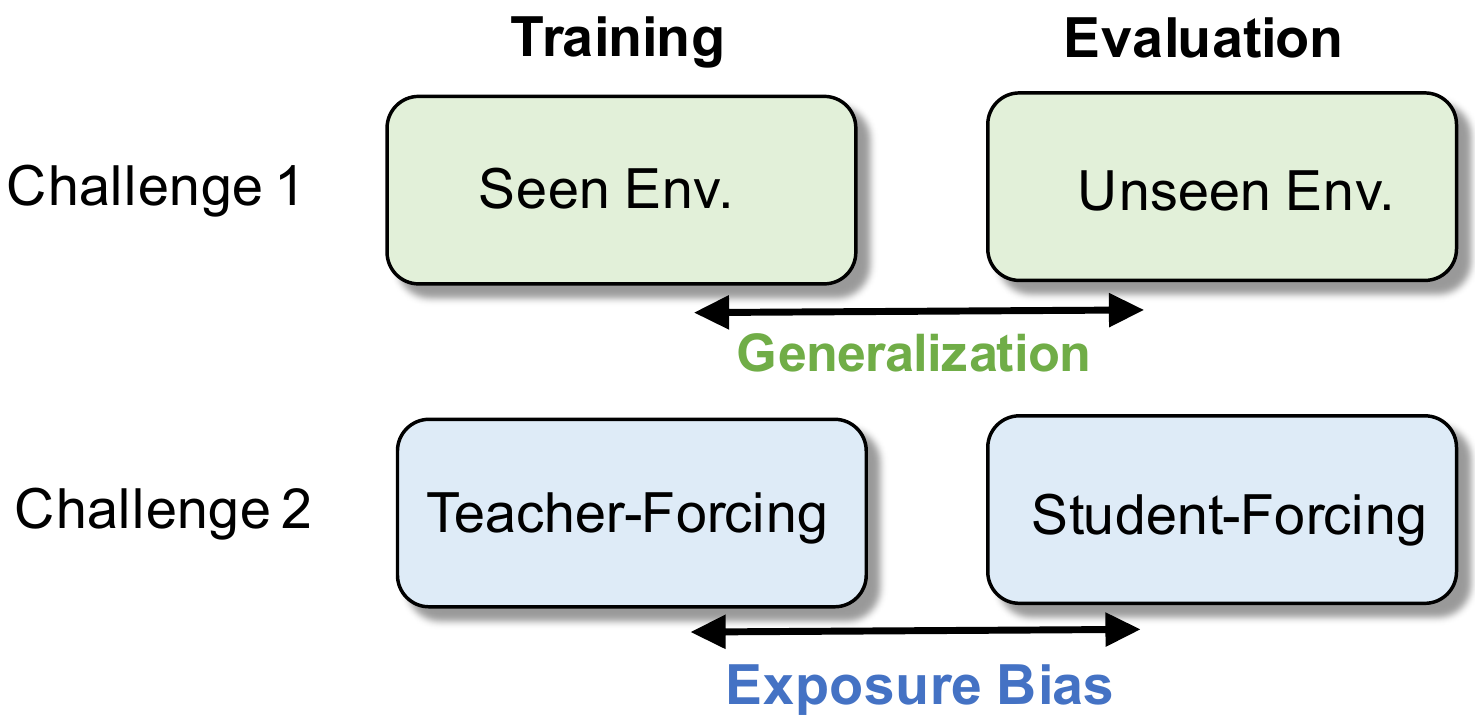}
	\end{tabular}
	\vspace{-3mm}
	\caption{Two challenges in VLN.}
	\vspace{-3mm}
	\label{fig:vln_challenge}
\end{figure}


In the recent Room-to-Room (R2R) VLN challenge~\cite{anderson2018vision}, most state-of-the-art methods are developed based on an encoder-decoder framework~\cite{cho2014learning,sutskever2014sequence}, where a natural language instruction is represented as a sequence of words, and a navigation trajectory as a sequence of actions, enhanced with attention~\cite{anderson2018vision,wang2018reinforced,fried2018speaker,ma2019self}. Two important components are shared by all VLN agents: 
(i) an {\em Instruction Encoder} that employs a language model (LM) for instruction understanding;  and
(ii) an {\em Action Decoder}, where an appropriate sequence-level training scheme is required for sequential decision-making. Each component faces its own challenges (see Figure~\ref{fig:vln_challenge}). 

The first challenge is generalizing grounded natural language instruction understanding from seen to unseen environments. Specifically, in the R2R task, only 69\% of bigrams are shared between training and evaluation.\footnote{Table~\ref{tab:ngram_stat} shows $n$-gram overlap statistics between training seen and validation seen/unseen environments.} Existing work leverages pretrained GloVe embeddings~\cite{pennington2014glove} to help generalize. In computer vision, it has been shown that large-scale models pretrained on ImageNet can transfer the knowledge to downstream applications~\cite{Yosinski2014}, thus improving generalization. Comparable language-based transfer learning has not been shown for instruction understanding in VLN.




\begin{table}[ht!]
\centering
\footnotesize
\begin{tabular}{@{}cc@{}|c@{}|c@{}}
\toprule
& \multicolumn{1}{c}{n-gram(s)} & \multicolumn{1}{c}{Validation Seen} & \multicolumn{1}{c@{}}{Validation Unseen} \\ 
\midrule
\multirow{4}{*}{\rotatebox{90}{\footnotesize Training}}& \texttt{1} & 87.2\% & 80.7\% \\
& \texttt{2} & 77.4\% & 68.9\% \\
& \texttt{3} & 65.6\% & 57.3\% \\
& \texttt{4} & 50.8\% & 44.4\% \\
\bottomrule
\end{tabular}
\caption{N-grams instruction overlap statistics between validation seen and unseen environments.}
\label{tab:ngram_stat}
\end{table}

The second challenge is \textit{exposure bias}~\cite{ranzato2015sequence} for the action decoder, due to the discrepancy between training and inference. This problem is common to many tasks where decoding is needed, including text generation, abstractive summarization, and machine translation~\cite{bengio2015scheduled}. 
Two widely used training strategies are \textit{student-forcing} and \textit{teacher-forcing} (described in detail in Section~\ref{sec:forcing}).
It is well-known that the sequence length determines which training strategy is more effective.
In the VLN literature, student-forcing has been widely used, as early work~\cite{anderson2018vision} used long trajectories (up to 20 steps) with a simple discrete action space.
Most recent work, however, has relied on a panoramic action space~\cite{fried2018speaker} in which most trajectories are only up to seven steps long. 
In such cases, teacher-forcing is preferable~\cite{tan2019learning}. 
Neither strategy is perfect: teacher-forcing has exposure bias, while student-forcing's random actions can cause an agent to deviate far from the correct path, rendering the original instruction invalid.\footnote{To compensate, beam search is often used to improve success rates. Recent work, e.g., using search strategies \cite{ke2019tactical} or progress monitors \cite{ma2019regretful}, has focused on mitigating the cost of computing top-$k$ rollouts.}

To tackle these challenges, we have developed two techniques to enable the agent to navigate more efficiently. For the first challenge, we leverage the recent large-scale pretrained language models, BERT~\cite{devlin2018bert} and GPT~\cite{radford2018improving}, 
to improve the agent's robustness in unseen environments. 
We show that large-scale \emph{language-only} pretraining improves generalization in grounded environments. 
%
For the second challenge, we propose a stochastic sampling scheme to balance teacher-forcing and student-forcing during training, so that the agent can recover from its own mistakes at inference time. 
As a result of combining both techniques, on the R2R benchmark test set, our agent (\short{})\footnote{\longname{}} achieves 53\% on SPL, 
an absolute 6\% gain over the current state of the art.

\section{Method}
\begin{figure}[t!]
	\vspace{-1mm}
	\centering
	\hspace{-2mm}
	\includegraphics[width=1.0\columnwidth]{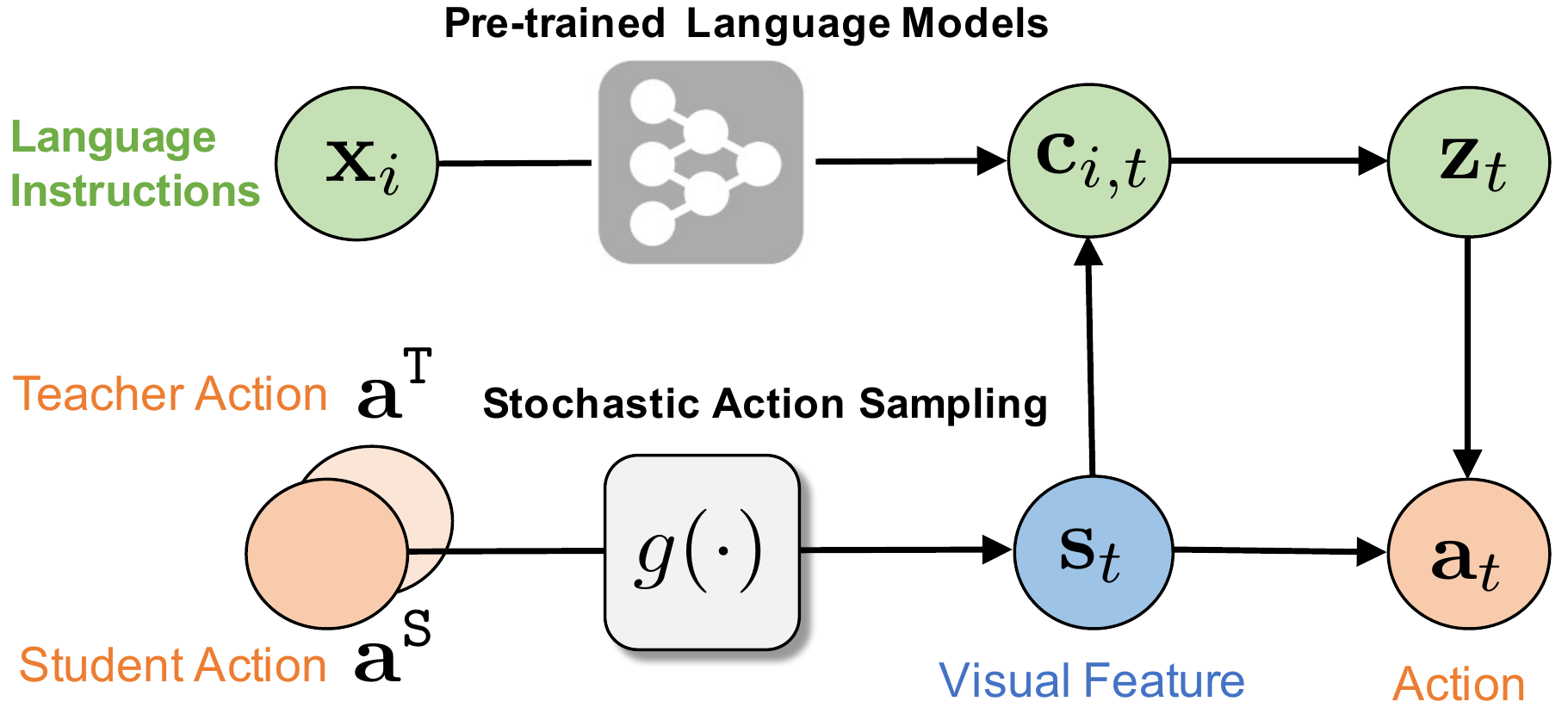}
	\vspace{-2mm}
	\caption{Illustration of proposed methods.}
	\vspace{-3mm}
	\label{fig:arch}
\end{figure}

In the VLN task, instructions are represented as a set $\Xcal=\{\xv_i\}_{i=1}^M$ of $M$ instructions per trajectory. Each instruction $\xv_i$ is a sequence of $L_i$ words, $\xv_i = [x_{i,1}, x_{i,2}, ..., x_{i,L_i}]$. Given $\Xcal$, the goal is to train an agent to navigate from a starting position $\sv_0$ to a target position, via completing a $T$-step trajectory $\tau = [\sv_0, \av_0, \sv_1, \av_1, \cdots, \sv_T, \av_T ] $, where $\sv_t$ and $\av_t$ are the visual state and navigation action, respectively, at step $t$.   
The training dataset $\Dcal_E = \{\tauv, \Xcal\}$ consists of example pairs of instruction set $\Xcal$ and a corresponding expert trajectory $\tauv$. Our goal is to learn a policy $\pi_{\thetav}(\tauv | \Xcal)$ that maximizes the log-likelihood of the target trajectory $\tauv$ given instructions $\Xcal$:
\vspace*{-1mm}
\begin{align}
\log \pi_{\thetav}(\tauv | \Xcal)  =\sum_{t=1}^{T}  \log \pi_{\thetav}(\av_t | \sv_t, \Xcal), 
\label{eq_factorization}
\end{align}
where $\thetav$ are trainable parameters.
The policy is usually parameterized as an attention-based seq2seq model, with a language encoder $\zv_t = f_{\theta_{E}} (\xv)$, and an action decoder $\av_t = f_{\theta_{D}} (\zv_t, \sv_t)$.
Successful navigation depends on 
$(\RN{1})$ precisely grounding the instructions $\Xcal$ in $\tau$ in various environments, and
$(\RN{2})$ correctly making the current decision $\av_t$  based on previous actions/observations $\tau_{< t} = [\sv_0, \av_0, \cdots, \sv_{t-1} ]$.
To address these concerns, we propose \short{}, illustrated in Figure~\ref{fig:arch}. 

\subsection{Instruction Understanding with Pretrained Language Models}

At each step $t$, the agent decides where to navigate by updating a dynamic understanding of the instructions $\zv_t$, according to its current visual state $\sv_{t}$. Given instruction $\xv$, the language encoder proceeds in two steps, end-to-end, by considering a function decomposition $f_{\theta_{E}} = f_{\thetav_{x\rightarrow e}} \circ  f_{\thetav_{e\rightarrow z}}$:  
%
\begin{itemize}
    \item $f_{\thetav_{x\rightarrow e}}: \xv \rightarrow  \ev$, where $\xv = [x_1, \cdots, x_L]$ is represented as its (contextualized) word embedding form $\ev = [e_1, \cdots, e_L]$, with $e_i$ as the representation for word $x_i$; 
    \vspace{-2mm}
    \item $f_{\thetav_{e\rightarrow z}}: \ev \rightarrow  \zv_t$: 
    For each embedded instruction $\ev$, we ground its representations as $\cv_{i,t}$ for state $\sv_t$ via neural attention. To handle language variability, one may aggregate features of multiple instructions $\mathcal{C}_t = \{\cv_{i,t} \}_{i=1}^M$ into a single joint feature~$ \zv_t = \frac{1}{M} \sum_{i=1}^M\cv_{i,t} $.\footnote{This recovers $\zv_t = \cv_t$ when only a single instruction is available. 
    }
    \vspace{-2mm}
\end{itemize}
%

Previous methods in VLN learn $\ev$ either from pretrained word embeddings~\cite{pennington2014glove} which do not take into account word context, or from scratch. 
As a result, their representations do not capture contextual information within each instruction. More importantly, they tend to overfit the training instructions associated with seen environments, limiting their utility in unseen environments. 
To remedy these issues, we propose to represent $\ev$ with contextualized word embeddings produced using large-scale pretrained language models, such as BERT and GPT.

\paragraph{Instruction Encoder.}
The agent's memory vector $\hv_{t-1}$ captures the perception and action history 
and is used to attend to the instruction $\xv$. A pretrained LM $f_{\thetav_{x\rightarrow e}}$ encodes the instruction $e = [\ev_1, \cdots, \ev_L]$; $\ev_i$ where the representation for word $x_i$, is built with $f_{\thetav_{x\rightarrow e}} \in \{$ \gpt{}, \bertbase{} $\}$, and $\thetav_{x\rightarrow e}$ are fine-tuned parameters. 
The embedded words $e = [\ev_1, \cdots, \ev_L]$ are passed through an LSTM $f_{\thetav_{e\rightarrow z}}$ to produce a sequence of textual features $[\hv^e_1, \cdots, \hv^e_L ]$. At each time step $t$, the {\it textual context} for the instruction $\xv$ is computed as weighted sum of textual features in the sequence:
\begin{align}
\cv_{i,t} = \sum_{l=1}^L \alpha_l \hv^e_l
\label{eq_instruction}
\end{align}
where $\alpha_l=\text{Softmax}(\hv_t^{\top} \hv^e_l )$, $\alpha_l$ places more weight on the word representations that are most relevant to the agent's current status.


\paragraph{Decoder.}
At each step, the agent takes an action $a_t$, and the environment returns new visual observations; the agent first performs one-hop visual attention $f(\cdot)$ to all the visual image features $s_t$, based on its previous memory vector $\hv_{t-1}$. Then, the agent updates its visual state $\sv_t$ as the weighted sum of the panoramic features, $\sv_t = \sum_j \gamma_{t,j} \sv_{t,j}$. The attention weight $\gamma_{t,j}$ for the $j$-th visual feature $\sv_{t,j}$ represents its importance with respect to the previous history context $\hv_{t-1}$, computed as 
$ \gamma_{t,j}= \mbox{Softmax} ( (\Wmat_h \hv_{t-1}  )^{\top} \Wmat_s \sv_{t, j})$~\cite{fried2018speaker} 
where 
$\mbox{Softmax} ( r_{j} ) = \exp( r_{j})/\sum_{j'} \exp( r_{j'})$, $\Wmat_h$ and $\Wmat_s$ are trainable projection matrices. 
\begin{align}
\hv_t  = f_{\thetav_D}([\sv_t, \av_{t-1}], \hv_{t-1})
\label{eq_history}
\end{align}
where $\av_{t-1}$ is the action taken at previous step, and $\thetav_{D}$ are the LSTM decoder parameters.

\paragraph{Two-stage learning.} The parameters of our agent are $\thetav = \{\thetav_{x\rightarrow e}, \thetav_{e\rightarrow z}, \theta_{D} \}$. In practice,  we find that the agent overfits quickly, when the full model is naively fine-tuned, with $\thetav_{x\rightarrow e}$ initialized by pretrained LMs (\eg BERT).
In this paper, we consider a two-stage learning scheme to facilitate the use of pretrained LMs for VLN.
$(\RN{1})$ Embedding-based stage: We fix $\thetav_{x\rightarrow e}$, and use BERT or GPT to provide instruction embeddings. Only $\{\thetav_{e\rightarrow z}, \thetav_{D} \}$ are updated (while tuning on validation). 
$(\RN{2})$ Fine-tuning stage: We train all model parameters $\thetav$ with a smaller learning rate, so that $\thetav_{x\rightarrow e}$ can adapt to our VLN task.

\subsection{Stochastic Action Sampling} \label{sec:forcing}
A core question is how to learn useful state representations $\sv_t$ in~Eq.~\eqref{eq_factorization} during the trajectory roll-out. In other words, which action should we use to interact with the environment to elicit the next state? As noted, most existing work uses one of two schemes:
$(\RN{1})$ {\em Teacher-forcing (TF)}, where the agent takes ground-truth actions $\av^{\mathtt{T}}$ only. Though TF enables efficient training, it results in ``exposure bias'' because agents must follow learned rather than gold trajectories at test time. In contrast, $(\RN{2})$ {\em Student-forcing (SF)}, where an action $\av^{\mathtt{S}}$ is drawn from the current learned policy, allows the agent to learn from its own actions (aligning training and evaluation), however, it is inefficient, as the agent explores randomly when confused or in the early stages of training.


In this work, we consider a stochastic scheme ({\em SS}) to alternate between choosing actions from $\av^{\mathtt{T}} $ and $\av^{\mathtt{S}}$ for state transition $ \sv \leftarrow g(\av^{\mathtt{T}}, \av^{\mathtt{S}} )$, inspired by scheduled sampling~\cite{bengio2015scheduled}. As illustrated in Figure~\ref{fig:arch}, at each step, the agent ``flips a coin'' with some probability $\epsilon$ to decide whether to take the teacher's action $a^{\mathtt{T}}$ or a sampled one $a^{\mathtt{S}}$:
\begin{align}
\av = \delta \av^{\mathtt{T}} + (1-\delta) \av^{\mathtt{S}},  
\label{eq_ss}
\end{align}
where $ \delta \sim \text{Bernoulli}(\epsilon)$. This allows the agent to leverage the advantages of both TF and SF, yielding a faster and less biased learner. We fix $\epsilon$ as a constant during learning, which is different from the decaying schedule in~\cite{bengio2015scheduled}. 


\section{Experiments}

\subsection{Dataset}
\label{sec:dataset}
We use the Room-to-Room dataset for the VLN task, built upon the Matterport3D dataset~\cite{chang2017matterport3d}, which consists of 10,800 panoramic views and 7,189 trajectories. Each trajectory is paired with three natural language instructions. The R2R dataset consists of four splits: train seen, validation seen, validation unseen, and test unseen. There is no overlap between seen and unseen environments. 
%
At the beginning of each episode, the agent starts at a specific location, and is given natural instructions, the goal of the agent is to navigate to the target location as quickly as possible. %

\subsection{Baseline Systems}
\label{sec:baselines}
We compare our approach with eight recently published systems:
\begin{itemize}[noitemsep,topsep=0pt]
\item \textsc{Random}: an agent that randomly selects a direction and moves five step in that direction ~\cite{anderson2018vision}. 
\item \textsc{Seq2Seq}: sequence-to-sequence model proposed by Anderson {\em et al.} as a baseline for the R2R benchmark~\cite{anderson2018vision} and analyzed in \cite{Thomason:19}.
\item \textsc{RPA}~\cite{wang2018look}: is an agent which combines model-free and model-based reinforcement learning, using a look-ahead module for planning. 
\item \textsc{Speaker-Follower}~\cite{fried2018speaker}: an agent trained with data augmentation from a speaker model with panoramic actions.
\item \textsc{Smna}~\cite{ma2019self}: an agent trained with a visual-textual co-grounding module and progress monitor on panoramic actions.
\item \textsc{RCM+SIL(train)}~\cite{wang2018reinforced}: an agent trained with cross-modal grounding locally and globally via reinforcement learning.
\item \textsc{Regretful}~\cite{ma2019regretful}: an agent with a trained progress monitor heuristic for search that enables backtracking. 
\item \textsc{Fast}~\cite{ke2019tactical}: an agent which combines global and local knowledge to compare partial trajectories of different lengths, enabling efficient backtrack after a mistake.
\item \textsc{EnvDrop}~\cite{tan2019learning}: proposed an environment dropout method, which can generate more environments based on the limited seen environments. 
\end{itemize} 

\subsection{Evaluation Metrics}
\label{sec:metrics}
We benchmark our agent on the following metrics:
\begin{itemize}[noitemsep,topsep=2pt] 
\item[\textbf{\texttt{TL}}] \textbf{Trajectory Length} measures the average length of the navigation trajectory.
\item[\textbf{\texttt{NE}}] \textbf{Navigation Error} is the mean of the shortest path distance in meters between the agent's final location and the target location. 
\item[\textbf{\texttt{SR}}] \textbf{Success Rate} with which the agent's final location is less than 3 meters from the target.
\item[\textbf{\texttt{SPL}}] \textbf{Success weighted by Path Length} trades-off \texttt{SR} against \texttt{TL}. 
\end{itemize}
\texttt{SPL} is the recommended primary metric, other metrics are considered as auxiliary measures.


\begin{table}[t]
\centering
\vspace{-2mm}
\small
\begin{tabular}{@{}c@{\hspace{0.5em}}c@{\hspace{0.5em}}|ll@{}|ll@{}}
\toprule
\multicolumn{2}{c}{} & \multicolumn{2}{c}{Validation Seen} & \multicolumn{2}{c@{}}{Validation Unseen} \\ 
Setting & Agent & \texttt{SR} $\uparrow$ & \hspace{-2mm}\texttt{SPL}  $\uparrow$ & \texttt{SR} $\uparrow$ & \hspace{-2mm}\texttt{SPL} $\uparrow$\\ 
\midrule
\multirow{2}{*}{S}
& seq2seq\phantom{+} & 51 & 46 & 32 & 25 \\
& \short & 47 (\textcolor{red}{\textbf{-4}}) & 43 (\textcolor{red}{\textbf{-3}}) & 43 (\textcolor{blue}{\textbf{+11}}) & 38 (\textcolor{blue}{\textbf{+13}})\\
\midrule
\multirow{2}{*}{M}
& seq2seq\phantom{+} & 49 & 44 & 33 & 26 \\
& \short & 56 (\textcolor{blue}{\textbf{+7}}) & 53 (\textcolor{blue}{\textbf{+9}}) & 56 (\textcolor{blue}{\textbf{+23}}) & 50 (\textcolor{blue}{\textbf{+24}}) \\
\bottomrule
\end{tabular}
\vspace{-1mm}
\caption{Comparison of \short{} and seq2seq.}
\vspace{-3mm}
\label{tab:greedy_res}
\end{table}

\subsection{Implementation}
\label{sec:imp}
We use a LSTM/\gpt{}/\bertbase{} for the language encoder, and a second single-layer LSTM for the action decoder (h=1024). We use Adamax and batch sizes of 24/16 for pretraining/finetuning. The learning rates for MLE are $1e^{-4}$, during finetuning \bertbase{} the learning rate is $5e^{-5}$. Following~\cite{fried2018speaker}, we use a panoramic action space and the ResNet image features provided by~\cite{anderson2018vision}. The code is publicly available here: \url{https://github.com/xjli/r2r_vln}.


\begin{table*}[ht!]
\small
\centering
\begin{tabular}{@{\hspace{3pt}}l@{\hspace{3pt}}l@{}r@{\hspace{9pt}}c@{\hspace{9pt}}c@{\hspace{9pt}}c|r@{\hspace{9pt}}c@{\hspace{9pt}}c@{\hspace{9pt}}c|r@{\hspace{9pt}}c@{\hspace{9pt}}c@{\hspace{9pt}}c}
\toprule
& & \multicolumn{4}{c}{Validation Seen} & \multicolumn{4}{c}{Validation Unseen} & \multicolumn{4}{c}{Test Unseen} \\ 
& Model & \texttt{TL} $\downarrow$ & \texttt{NE} $\downarrow$ & \texttt{SR} $\uparrow$ & \texttt{SPL} $\uparrow$ & \texttt{TL} $\downarrow$  & \texttt{NE} $\downarrow$ & \texttt{SR} $\uparrow$ & \texttt{SPL} $\uparrow$ & \texttt{TL} $\downarrow$ & \texttt{NE} $\downarrow$ & \texttt{SR} $\uparrow$ & \texttt{SPL} $\uparrow$\\ 
\midrule
& \textsc{Random} & 9.58 & 9.45 & 16 & - & 9.77 & 9.23 & 16 & - & 9.93 & 9.77 & 13 & 12 \\
& \textsc{Seq2Seq} & 11.33 & 6.01 & 39 & - & 8.39 & 7.81 & 22 & - & \phantom{0,0}8.13 & 7.85 & 20 & 18 \\
& \textsc{RPA} & - & 5.56 & 43 & - & - & 7.65 & 25 & - & \phantom{0,0}9.15& 7.53 & 25 & 23 \\
\multirow{2}{*}{\rotatebox{90}{\footnotesize Greedy}} & \textsc{Speaker-Follower} & - & 3.36 & 66 & - & - & 6.62 & 35 & - & \phantom{0,0}14.82 & 6.62 & 35 & 28\\
& \textsc{SMNA} & - & - & - & - & - & - & - & - & \phantom{0,0}18.04 & 5.67 & 48 & 35 \\
& \textsc{RCM+SIL(train)} & 10.65 & 3.53 & 67 & - & 11.46  & 6.09 & 43 & - & 11.97 & 6.12 & 43 & 38 \\
& \textsc{Regretful} & - & 3.23 & 69 & 63 & - & 5.32 & 50 & 41 & 13.69 & 5.69 & 48 & 40 \\
& \textsc{Fast} & - & - & - & - & 21.17 & 4.97 & 56 & 43 & 22.08 & 5.14 & 54 & 41 \\
& \textsc{EnvDrop} & 11.00 & 3.99 & 62 & 59 & 10.70 & 5.22 & 52 & 48 & 11.66 & 5.23 & 51 & 47 \\
\rowcolor{Gray}
\cellcolor{white} & \short{} & 10.35 & \textcolor{blue}{\textbf{3.09}} & \textcolor{blue}{\textbf{71}} & \textcolor{blue}{\textbf{67}} & 10.06 & \textcolor{blue}{\textbf{4.31}} & \textcolor{blue}{\textbf{59}} & \textcolor{blue}{\textbf{55}} & 10.52 & \textcolor{blue}{\textbf{4.53}} & \textcolor{blue}{\textbf{57}} & \textcolor{blue}{\textbf{53}} \\

\midrule
& Human & - & - & - & - & - & - & - & - & \phantom{0,0}11.85 & 1.61 & 86 & 76 \\
\bottomrule
\end{tabular}
\vspace{-1mm}
\caption{Comparison with the state-of-the-art methods. \textcolor{blue}{\textbf{Blue}} indicates best value overall. 
}
\label{tab:main_result}
\vspace{-2mm}
\end{table*}

\subsection{Results}
\label{sec:results}
\paragraph{Robust Generalization.} 
First, we compare \short{} to a baseline seq2seq model\footnote{The baseline seq2seq agent is the \textsc{Follower} of \textsc{Speaker-Follower}~\cite{fried2018speaker}.} in two evaluation settings on the validation splits: (1) \textbf{S}: A single instruction is provided to the agent at a time. Thus, three separate navigation trajectories are generated corresponding to three alternative instructions in this setting. We report the averaged performance over three separate runs. (2) \textbf{M}: All three instructions are provided to the agent at once. The seq2seq baseline does not have an aggregation strategy so we report its performance for the single trajectory with maximum likelihood. For \short{}, we aggregate the instructions via context mean-pooling and generate a single trajectory. No data augmentation is applied to either model.

The results are summarized in Table~\ref{tab:greedy_res}.
$(\RN{1})$
\short{} drastically outperforms the seq2seq models on unseen environments in both settings, and $(\RN{2})$ Interestingly, our method shows a much smaller gap between seen and unseen environments than seq2seq. It demonstrates the importance of pretrained LMs and stochastic sampling for strong generalization in unseen environments.

\paragraph{Comparison with SoTA.}
In Table~\ref{tab:main_result}, we compare the performance of our agent against all the published methods, 
our \short{} agent outperforms the existing models on nearly all the metrics. 
%
\paragraph{Ablation Analysis.}
Key to this work is leveraging large-scale pretrained LMs and effective training strategies for action sequence decoding. 
Table~\ref{tab:encoder_train_result} shows an ablation of these choices.
(1)
\bertbase{} and \gpt{} are better than LSTM on both seen and unseen environments, and \bertbase{} generalizes better than \gpt{} on unseen environments.  
(2) Teacher-forcing performs better than student-forcing on validation unseen environments, while an opposite conclusion is drawn on validation seen environments. 
SS performs the best on unseen environments.

\begin{table}[ht!]
\footnotesize
\centering
\begin{tabular}{@{\hspace{0pt}}c@{\hspace{2pt}}c|r@{\hspace{5pt}}c@{\hspace{5pt}}c@{\hspace{5pt}}c@{\hspace{3pt}}|@{\hspace{3pt}}r@{\hspace{5pt}}c@{\hspace{5pt}}c@{\hspace{5pt}}c@{}}
\toprule
\multicolumn{2}{c}{} & \multicolumn{4}{c}{Validation Seen} & \multicolumn{4}{c@{}}{Validation Unseen}  \\ 
LM & & \texttt{TL}  & \texttt{NE}  & \texttt{SR}  & \texttt{SPL}  & \texttt{TL}  & \texttt{NE}  & \texttt{SR} & \texttt{SPL} \\ 
\midrule
\multirow{3}{*}{\rotatebox{90}{LSTM}} & TF & 10.50 & 5.74 & 44 & 42 & \phantom{0,0}9.86 & 6.23 & 42 & 39 \\
& SF & 11.87 & 3.97 & 59 & 53 & \phantom{0,0}13.23 & 6.17 & 40 & 31 \\
& SS & 10.99 & 3.46 & 64 & 59 & \phantom{0,0}10.73 & 4.89 & 53 & 48 \\
\midrule
\multirow{3}{*}{\rotatebox{90}{\gpt}} & TF & 10.03 & 4.05 & 60 & 58 & \phantom{0,0}9.43 & 3.36 & 49 & 46 \\
 & SF & 11.46 & 2.53 & 73 & 67 & \phantom{0,0}13.13 & 5.13 & 49 & 41 \\
 & SS & 10.60 & 2.99 & 71 & 68 & \phantom{0,0}10.79 & 3.05 & 56 & 51 \\
\midrule
\multirow{3}{*}{\rotatebox{90}{\bertbase}} & TF & 10.57 & 4.06 & 59 & 56 & \phantom{0,0}9.61 & 5.13 & 51 & 47 \\
 & SF &  12.39 & 2.71 & 73 & 64 & \phantom{0,0}13.12 & 5.06 & 51 & 42 \\
 & SS &  10.35 & 3.09 & 71 & 67 & 10.06 & 4.31 & \textcolor{blue}{\textbf{59}} & \textcolor{blue}{\textbf{55}} \\
\bottomrule
\end{tabular}
\caption{Ablation results of different language pretrainings and training strategies: Teacher Forcing (TF), Student Forcing (SF) and Stochastic Sampling (SS).}
\label{tab:encoder_train_result}
\end{table}




\paragraph{Qualitative Examples.}
We provide two navigation examples of \short{} on the validation unseen environments with the step-by-step views and top-down views in Appendix.

(1) Figure~\ref{fig:lstm_bert} shows how the agent with LSTM instruction encoder performs compared with our \short{} agent. There are two rare words ``\textcolor{blue}{\textbf{mannequins}}'' and ``\textcolor{blue}{\textbf{manikins}}'' which are not in the training dataset and confuse the LSTM agent, while, \short{} successfully maps these two ``mannequins'' and ``manikins'' to the correct objects. 

(2) The second set in Figure~\ref{fig:training_exampls} shows how the agents trained with different training strategies performs in an unseen environment. The agents trained with teacher-forcing and student-forcing both fail, while \short{} succeeds. 



\section{Conclusion}
We present \short{}, a navigation agent based on two previously underexplored techniques in VLN: pretrained language models
and stochastic action sampling. Our \short{} demonstrates robust generalization in the unseen environments, leading to a new state-of-the-art performance over many of the much more complex approaches previously proposed. As both the components of \short{} can be easily integrated, future models can consider building upon them as a strong baseline system. 
\section*{Acknowledgments}
We thank the anonymous reviewers for their insightful comments, NSF IIS-1703166, DARPA's CwC program through ARO W911NF-15-1-0543, and the Allen Institute for Artificial Intelligence.

\bibliography{emnlp-ijcnlp-2019}
\bibliographystyle{acl_natbib}

\appendix

\begin{figure*}[ht!]
$\mathtt{Instruction~{\bf A}}$: {\em Go up the stairs to the right, turn left and go into the room on the left. Turn left and stop near the \textcolor{blue}{\textbf{mannequins}}.}\\
$\mathtt{Instruction~{\bf B}}$: {\em Walk up the small set of stairs.  Once you reach the top, turn 45 degrees to your left. Walk through the door at the bottom of the large staircase. After you are inside, turn left and wait near the statue.}
\\
$\mathtt{Instruction~{\bf C}}$: {\em Walk up the stairs Through the doorway on the left. Make a left in the room and stop before the two \textcolor{blue}{\textbf{manikins}}.} \\
\vspace{-1mm}
\begin{tabular}{c}
    \begin{subfigure}[t]{1.0\textwidth}
        \centering
        \includegraphics[trim={0cm 0cm 0cm 0cm},clip,scale=0.69]{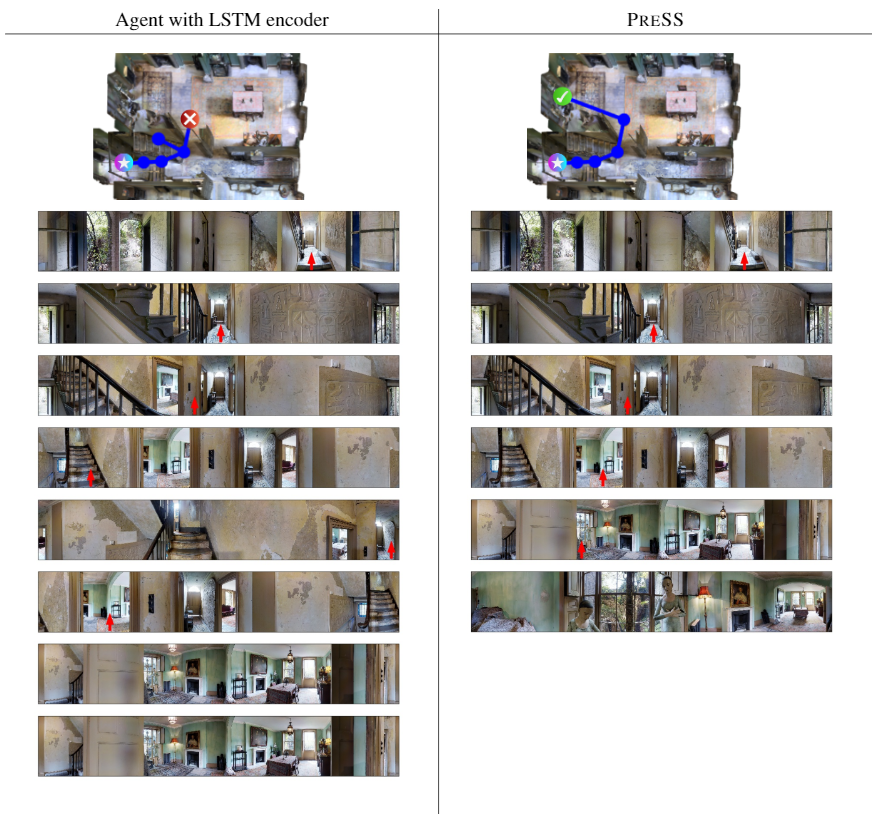}
        \label{fig:}
    \end{subfigure}
    \end{tabular}
    \raggedright
    \caption{Comparison between the agent equipped with an LSTM instruction encoder and our \short{} agent on a validation unseen environment (path\_id: 6632), including top-down trajectory view and step-by-step navigation views. We indicate the start 
    (\protect\inlinegraphics{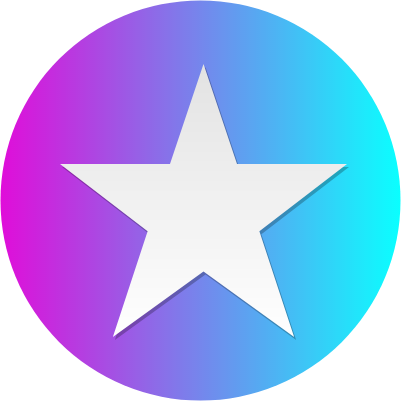}), target 
    (\protect\inlinegraphics{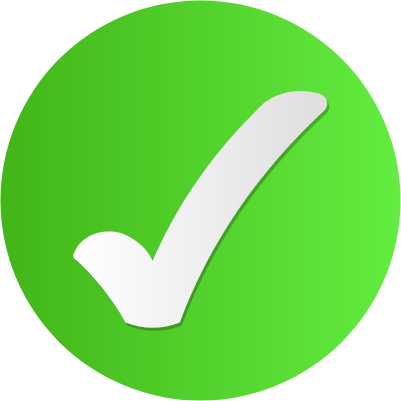}) and failure 
    (\protect\inlinegraphics{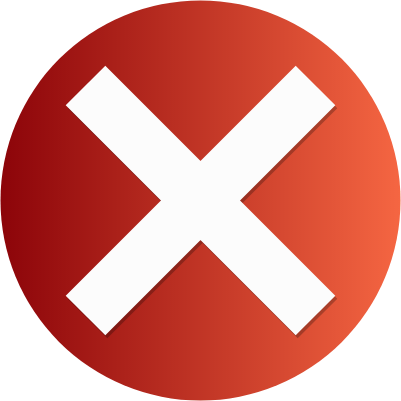}) of agents in an unseen environment.}
\label{fig:lstm_bert}
\end{figure*}

\begin{figure*}[ht!]
$\mathtt{Instruction~{\bf B}}$: {\em Walk up the stairs. Next, walk inside through the sliding glass doors. Continue straight past the television, towards another set of stairs.  Wait near the bottom of stairs.} \\
\vspace{-2mm}
\begin{tabular}{c}
    \begin{subfigure}[t]{1.0\textwidth}
        \centering
        \includegraphics[trim={0cm 0 0cm 0cm},clip,scale=0.69]{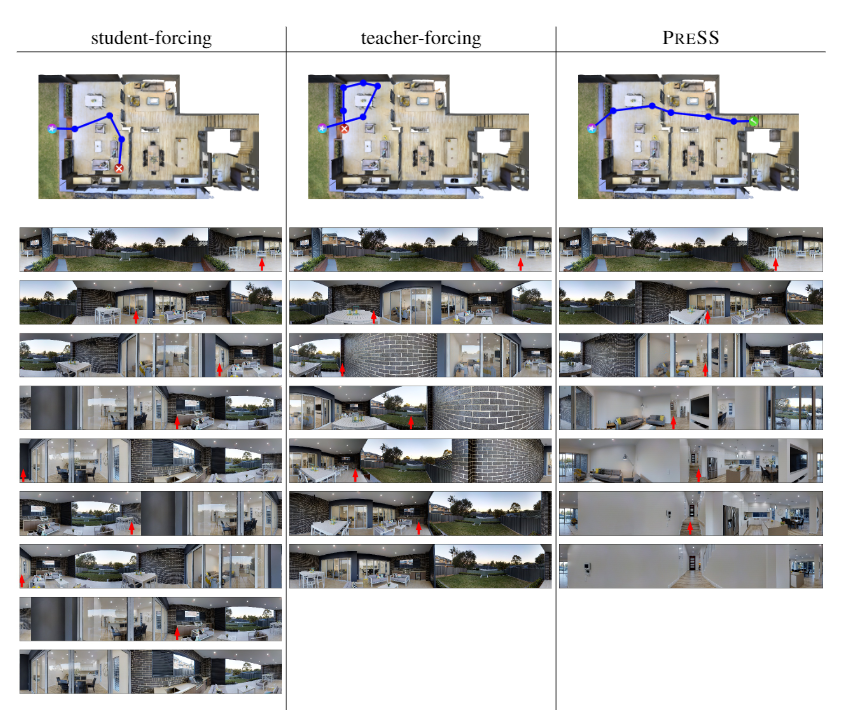}
        \label{fig:}
    \end{subfigure}
    \end{tabular}
    \raggedright
    \caption{Comparison among the agents trained with teacher-forcing, student-forcing and stochastic sampling strategies on a validation unseen environment (path\_id: 7201), including top-down trajectory view and step-by-step navigation views. We indicate the start 
    (\protect\inlinegraphics{./start-c.png}), target 
    (\protect\inlinegraphics{./stop-yes-c.png}) and failure 
    (\protect\inlinegraphics{./stop-no-c.png}) of agents in an unseen environment.}
\label{fig:training_exampls}
\end{figure*}

\end{document}